\documentclass[letterpaper, 10 pt, conference]{ieeeconf}  

\IEEEoverridecommandlockouts                              

\usepackage{graphics} 
\usepackage{epsfig} 
\usepackage{mathptmx} 
\usepackage{times} 
\usepackage{amsmath} 
\usepackage{amssymb}  
\usepackage{url}
\usepackage{subfig}
\usepackage{booktabs}
\usepackage{float} 
\usepackage{stfloats}

\makeatletter
\let\NAT@parse\undefined
\makeatother
\usepackage{hyperref}  

\title{\LARGE \bf 
Whleaper: A 10-DOF Flexible Bipedal Wheeled Robot}

\author{Yinglei Zhu$^{1\dag}$, Sixiao He$^{1\dag}$, Yan Ning$^{2}$, Zhenghao Qi$^{1}$, Zhuoyuan Yong$^{1}$, Yihua Qin$^{1}$ and Jianyu Chen$^{3*}$
\thanks{$^{\dag}$Contributed equally to this work.}
\thanks{$^{1}$are with Tsinghua University, Beijing, China.}%
\thanks{$^{2}$is with the Hong Kong University of Science and Technology, Hong Kong SAR, China.}%
\thanks{$^{3}$is with Institute for Interdisciplinary Information Sciences, Tsinghua University, Beijing, China.}%
\thanks{$^{*}$Corresponding author: Jianyu Chen, jianyuchen@tsinghua.edu.cn.}%
}

\begin{document}
\maketitle
\thispagestyle{empty}
\pagestyle{empty}


\begin{abstract}

Wheel-legged robots combine the advantages of both wheeled robots and legged robots, offering versatile locomotion capabilities with excellent stability on challenging terrains and high efficiency on flat surfaces. However, existing wheel-legged robots typically have limited hip joint mobility compared to humans, while hip joint plays a crucial role in locomotion. 
In this paper, we introduce Whleaper, a novel 10-degree-of-freedom (DOF) bipedal wheeled robot, with 3 DOFs at the hip of each leg. Its humanoid joint design enables adaptable motion in complex scenarios, ensuring stability and flexibility. This paper introduces the details of Whleaper, with a focus on innovative mechanical design, control algorithms and system implementation. 
Firstly, stability stems from the increased DOFs at the hip, which expand the range of possible postures and improve the robot's foot-ground contact. Secondly, the extra DOFs also augment its mobility. During walking or sliding, more complex movements can be adopted to execute obstacle avoidance tasks. Thirdly, we utilize two control algorithms to implement multimodal motion for walking and sliding.
By controlling specific DOFs of the robot, we conducted a series of simulations and practical experiments, demonstrating that a high-DOF hip joint design can effectively enhance the stability and flexibility of wheel-legged robots. Whleaper shows its capability to perform actions such as squatting, obstacle avoidance sliding, and rapid turning in real-world scenarios.

\end{abstract}


\section{INTRODUCTION}

Mobile robots can be divided into aerial and ground-based systems. Aerial robots are susceptible to changes in external airflow \cite{c1} and have limitations in endurance and positioning accuracy \cite{c2}. Ground robots are classified according to their mobility mechanisms as wheeled, legged, or crawler-based \cite{c3}. Wheeled robots are fast and efficient on flat terrain but struggle with obstacle traversal, terrain adaptability, and turning efficiency \cite{c4}, \cite{c5}. On the other hand, legged robots, particularly bipedal ones, demonstrate exceptional adaptability across diverse terrains and excel in navigating obstacles \cite{c6}, \cite{c7}, \cite{c8}, \cite{c9}. Legged robots provide improved flexibility, maneuverability, and stability \cite{c10}, \cite{c11}. Nevertheless, their movement speeds are typically lower, and they are prone to rollovers. Wheel-legged robots have attracted considerable research interest in recent years. They combing the strengths of wheeled robots with legged robots, enabling fast and agile locomotion on flat terrain while effectively navigating obstacles \cite{c12}. Wheel-legged robots combine the high efficiency of wheels with the terrain adaptability of legs. However, existing wheel-legged robots often have only one \cite{c15}, \cite{c16}, \cite{c17} or two \cite{c18}, \cite{wlr} DOFs at the hip joint. This limitation impedes tasks such as lateral walking and obstacle avoidance sliding. To enhance motion flexibility, it is imperative to increase the robot’s DOFs.

Our goal is to design a 10-DOF bipedal wheeled robot, Whleaper, with 3 DOFs at each hip, similar to humans. Using optimal control theory and reinforcement learning control, we have developed control methods for our robot to perform multimodal motion. Compared to other robots, including legged and wheeled ones, and existing bipedal wheeled robots such as Ascento from ETH Zurich \cite{c20}, our robot distinguishes itself with more DOFs at hip, a complex structure, and enhanced movement stability and flexibility.

\begin{figure}[thpb]
      \centering
      \includegraphics[width=0.95\linewidth]{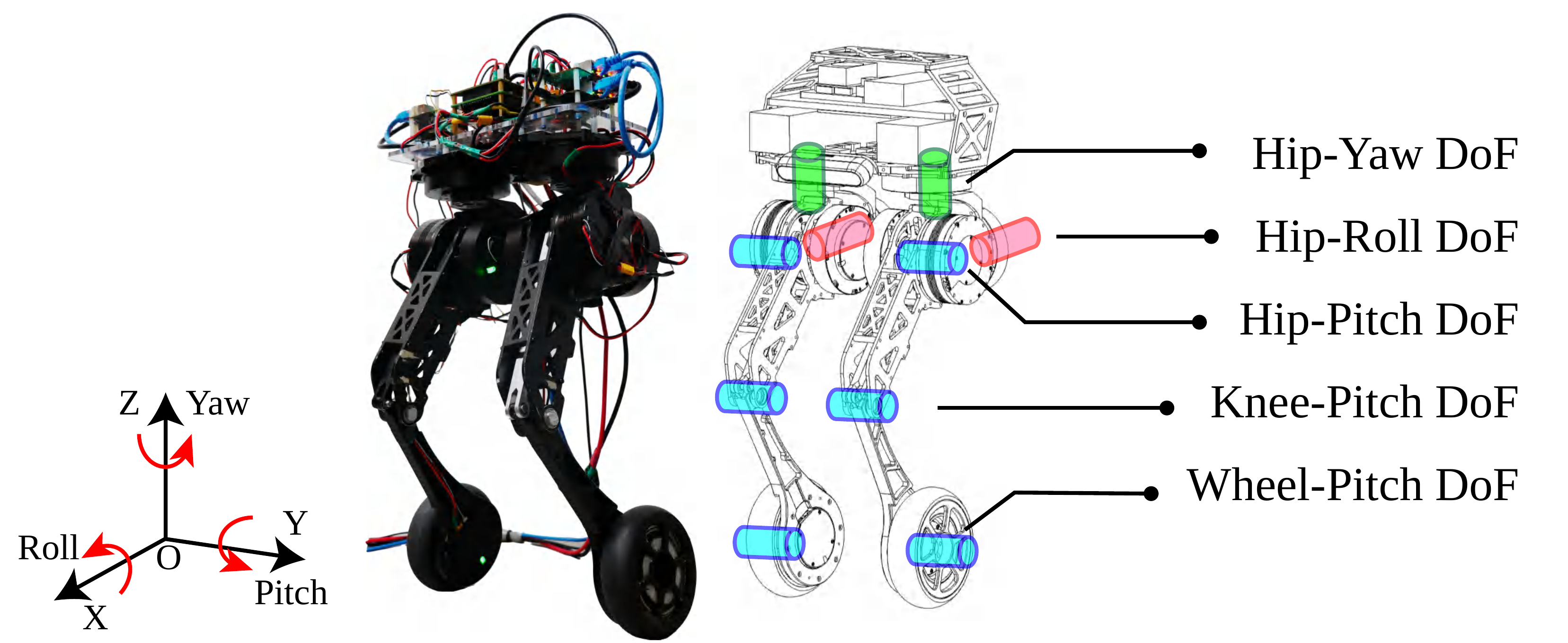}
      \caption{Prototype of Whleaper robot and schematic diagram of DOF configuration.}
      \label{fig: The actual prototype of Whleaper.}
\end{figure}

The main contributions of this paper can be summarized as follows: 

\begin{itemize}
    \item Mechanical design of a 10-DOF bipedal wheel-legged robot, Whleaper, with terrain adaptability and high-performance mobility. 
    \item Development of tailored control methods, including Linear Quadratic Regulator (LQR) and Reinforcement Learning (RL), which enable smooth transitions between sliding and walking modes.
    \item Establishment of the physical system for Whleaper, including both hardware and software.
    \item Validation of the mechanical design and algorithms via a series of simulations and real-world experiments.
\end{itemize}
The paper follows the structure outlined below:

\begin{itemize}
    \item Section \ref{section: system description} introduces the robot's system, including mechanical design, hardware, and software.
    \item Section \ref{section: algorithm} explains the dynamic model and control methodology, involving the LQR equation, balancing control, and reinforcement learning strategy.
    \item Section \ref{section: experiments} presents an overview of the experiment settings, results from simulations and real-world scenarios. 
    \item Finally, in section \ref{section: conclusions}, we conclude the paper and outline avenues for future work.
\end{itemize}


\section{SYSTEM DESCRIPTION}\label{section: system description}

\subsection{Mechanical Design}

\begin{figure}[thpb]
      \centering
    \includegraphics[width=0.9\linewidth]{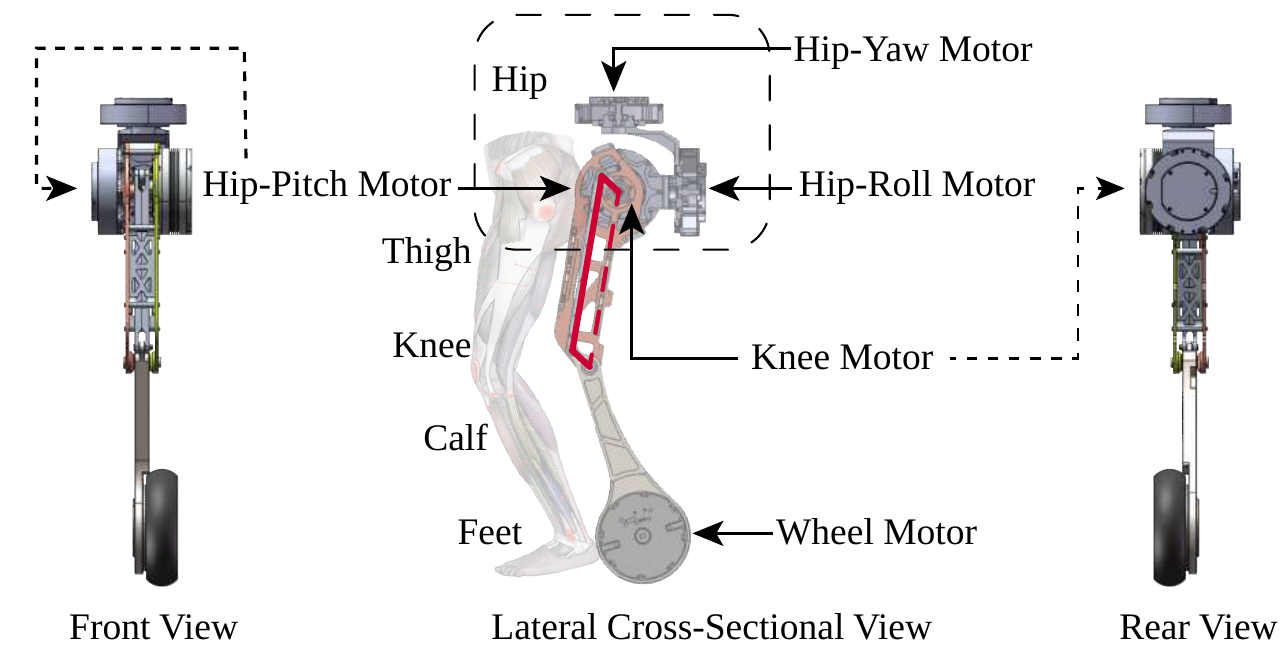}
      \caption{Schematic diagram of joint space.}
      \label{fig: joint space schematic diagram}
\end{figure}

Whleaper's legs have 5 DOFs each. As mentioned above, we aim to improve the stability and flexibility of the wheel-legged robot by increasing the DOFs in hip joints, and maintain its capabilities for both wheeled and legged locomotion. 

It is necessary to demonstrate the rationale for this increase in DOFs. From a biological perspective, the hip of a human has 3 DOFs to ensure smooth walking \cite{hip}. The distinctive difference between the wheel-legged robot and humans is the presence of wheels for ground contact. A well-designed system should allow the wheels to have a wide range of orientations relative to the ground, thereby providing a greater variety of foot placement options. 

Considering a wheel in contact with the ground, its orientation can be described using just two coordinates due to its axial symmetry (namely roll and yaw. Pitch represents the rotational direction of the wheel). Similar to ETH's Ascento and Tencent's Ollie \cite{ollie}, our design incorporates hip-pitch and knee-pitch motions for maintaining ground contact and improving adaptability to uneven terrain. Therefore, in our design, the orientation of the wheel is achieved by adding hip-roll and hip-yaw DOFs, while hip-pitch and knee-pitch DOFs ensure the walking motion.

The robot is composed of a base, hip joints, thighs, calves, and wheels. The hip joints are constructed with ball-wrist structures, providing 2 DOFs, yaw and roll, for thighs. Meanwhile, the hip joints’ ends and the thighs are connected by motors, which offering pitch DOFs to lift the robot’s legs.

Furthermore, we have implemented a series-parallel hybrid mechanical connection structure for the legs. The four-bar linkage reduces knee joint weight, lowers leg inertia, and enables precise leg control. Moreover, it efficiently transfers torque from the hip to the knee joint. So the overall leg structure is kinematically equivalent to a serial manipulator. This configuration is advantageous for both kinematic and dynamic solutions. At the end, a motor is directly attached to drive the rotation of wheels. 

From the practical perspective, the hip-yaw and hip-roll movements play a crucial role in establishing the grounding posture of the wheels, while the hip-pitch motion facilitates the lifting motion of the thigh. The knee motion ensures continuous contact between the wheels and the ground, fulfilling the required constraints.

\begin{table}[!ht]
    \centering
    \caption{Comparison of joint ranges (deg.) between Whleaper and the human body}
    \label{table: joint ranges}
    \begin{tabular}{ccc}
        \toprule 
        \textbf{Joint} & \textbf{Human Range} & \textbf{Robot Range } \\ 
        \midrule
        Hip Roll & -48,+31 & -20,+5  \\ 
        Hip Pitch & -113,+28 & -60,+0  \\ 
        Hip Yaw & -45,+45 & -5,+5  \\ 
        Knee Pitch & -0,+134 & -0,+120  \\ 
        \bottomrule
    \end{tabular}
\end{table}

Table \ref{table: joint ranges} illustrates that Whleaper has a human-like range of motion in the knee, with 3 DOFs in its hip, but its overall range of motion still falls short of that of humans.


\subsection{Hardware}

\begin{figure*}[!thbp]
      \centering
      \includegraphics[width=0.9\linewidth]{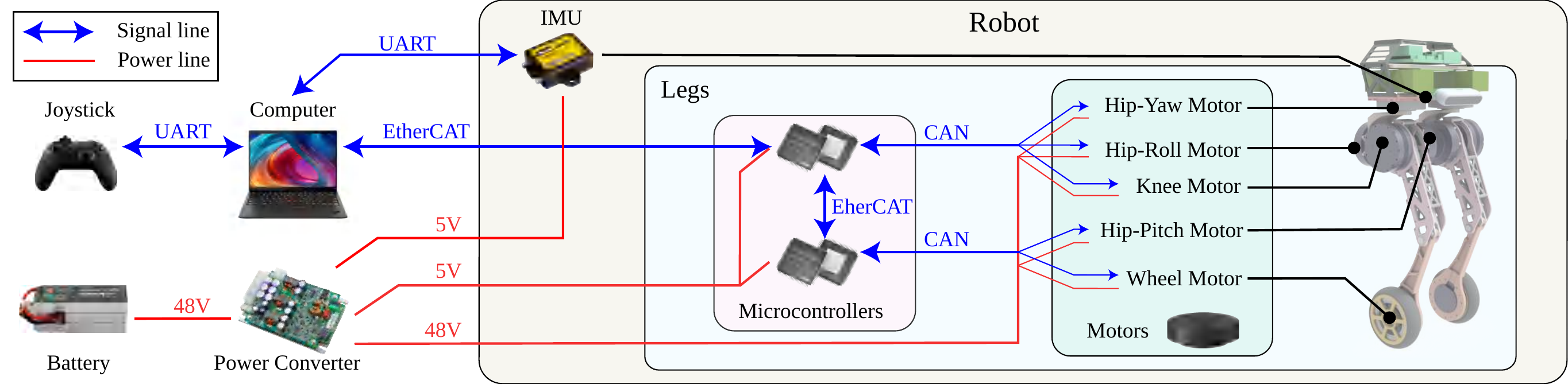}
      \caption{Hardware connection diagram.}
      \label{fig: hardware connection diagram}
\end{figure*}

Tailored to its unique mechanical structure, we have designed a specialized electrical and communication system for Whleaper, as depicted in Figure \ref{fig: hardware connection diagram}.

Because of concentrated motors and aluminum structure at the base and hip, Whleaper has a high center of gravity (COG). We use THU-8108 \cite{motors} motors for efficient control of multiple joints, delivering peak torques of 18 Nm. Specifically, THU-8112 \cite{motors} motors with peak torques of 50 Nm are chosen for the calves to withstand significant knee forces. Each motor is equipped with dual encoders for precise feedback on position, velocity, and torque, adaptable to closed-loop control methods.

We use the MicroStrain 3DM-GX5-AHRS IMU for high-precision state capture, featuring 1000 Hz sampling.

In terms of communication, two microcontrollers serve as bridges between the motors and the computer. To enhance real-time performance, a hybrid communication approach is employed between the computer, microcontrollers, and motors, utilizing both Ethernet for Control Automation Technology (EtherCAT) and Controller Area Network (CAN).

For a stable power supply, we use a 48V lithium-ion polymer (LiPo) battery pack. 

Finally, a list of components can be found in Table \ref{table: component list}.

\begin{table}[!thpb]
    \caption{Components list}
    \label{table: component list}
    \begin{center}
    \begin{tabular}{|l|l|l|}
    \hline
        \textbf{Component} & \textbf{Name} \\ \hline
        Computer & HP OMEN 16 GAMING LAPTOP \\ \hline
        Knee Motors & THU-A8112-09\\ \hline
        Other Motors & THU-A8108-06 \\ \hline
        Microcontrollers & XMC4300-F100F256 \\ \hline
        IMU & MicroStrain 3DM-GX5-AHRS \\ \hline
        Joystick & Xbox Elite Wireless Controller Series 2 \\ \hline
        Battery & GS-5300mAh-30C-6S1P \\ \hline
        Miniature Circuit Breaker & SIEMENS 5SJ52 \\ \hline
    \end{tabular}
    \end{center}
\end{table}
\subsection{Software}

\begin{figure}[!hpb]
      \centering
      \includegraphics[width=\linewidth]{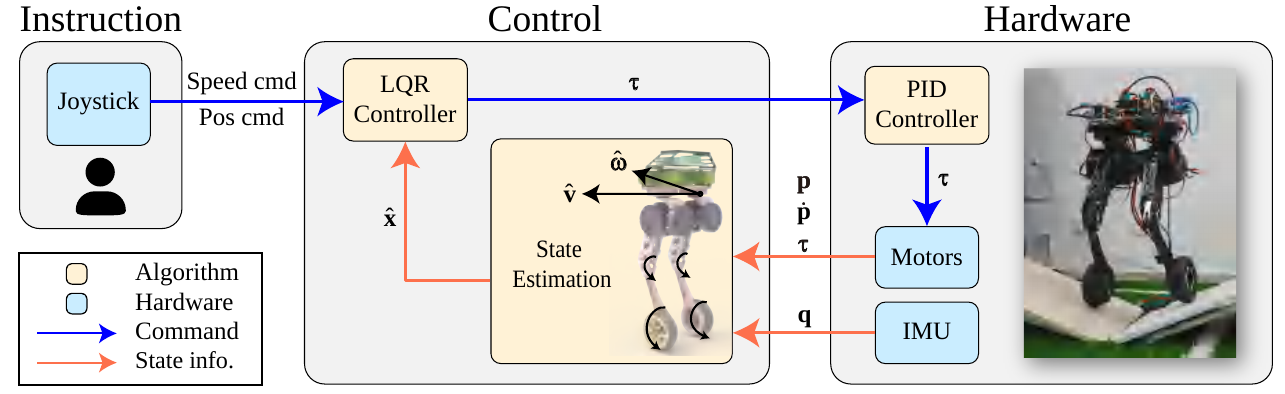}
      \caption{Control system architecture.}
      \label{fig: control architecture}
\end{figure}

As depicted in Figure \ref{fig: control architecture}, the software architecture deployed on Whleaper is structured into three layers: the instruction layer, the control layer, and the hardware layer. 

The instruction layer is responsible for user interaction and command transmission. Users can input desired pose and speed into the control layer by manipulating a joystick, while simultaneously observing real-time feedback of overall posture and joint states through the user interface (UI). 

As a critical link in feedback control, the control layer is tasked with processing information and issuing commands. It calculates the estimated state vector by integrating data from the IMU and motors. Subsequently, the designated controller, such as LQR, computes motor torque commands. These commands are then transmitted to the hardware layer, and smoothly adjust the robot to the desired state.

The hardware layer is responsible for low-level data acquisition and control of hardware components including the IMU and motors. At the motor’s low-level control, a PID controller is employed. The hardware layer also forwards overall attitude information from the IMU and joint states from motor encoders to the control layer. 


\section{ALGORITHM}\label{section: algorithm}

In order to enable multiple modes of locomotion, namely sliding and walking, we have developed dedicated control algorithms for each. We utilize an optimal control algorithm (LQR) to achieve balanced sliding, while a reinforcement learning method based on Proximal Policy Optimization (PPO) \cite{PPO} is employed to facilitate walking, jumping, etc.

\subsection{Sliding: LQR Control for Balancing}

\subsubsection{Coordinate System and Symbols}
\begin{figure}[htbp]
\centering
\subfloat[]{\includegraphics[width=0.4\linewidth]{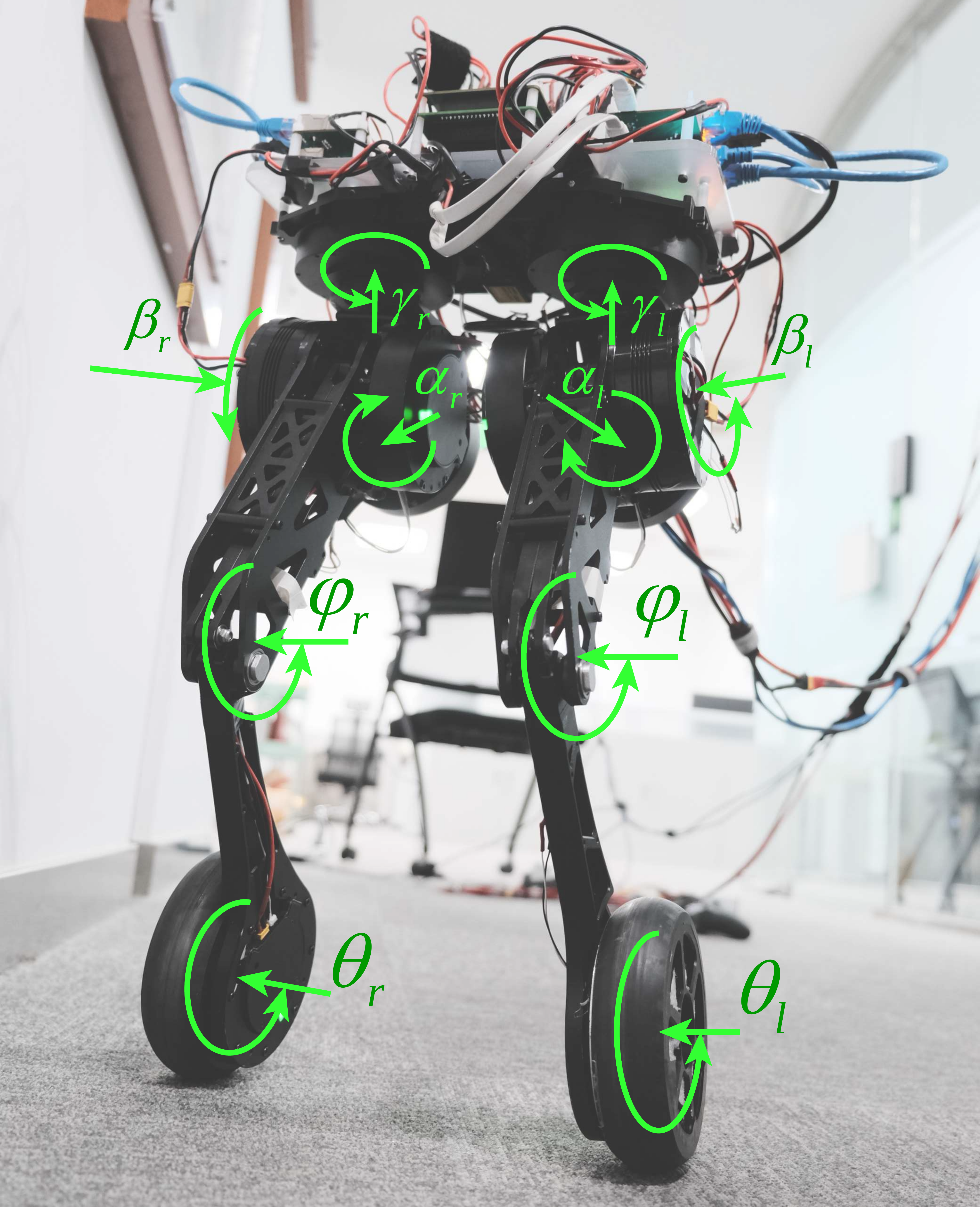}}
\hfill
\subfloat[]{\includegraphics[width=0.4\linewidth]{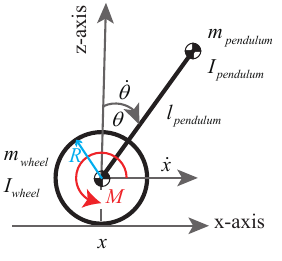}}
\caption{Schematic representation of dynamics model.}
\label{fig: LQR}
\end{figure}

The coordinate system is established as depicted in Figure \ref{fig: LQR}. 

We designate the rotations along the roll, pitch, and yaw axes of the hip as $\alpha$, $\beta$, and $\gamma$ respectively. The rotation of the knee is denoted by $\phi$, and $\theta_{r/l}$ represents the rotation of wheel, while the subscripts \textit{r} and \textit{l} represent right and left. Considering the design specifications of the mechanism, where the range of motion is limited, we assume a momentarily constant mass for the base, the hip, and the connected thigh. Additionally, considering that the angles $\alpha$ and $\gamma$ fall within a narrow range, we make the assumption that the leg can be effectively approximated as a planar rigid body in motion. We abstract the system as a wheeled inverted pendulum system (WIPS) \cite{kim2011dynamic} and establish the coordinate system as shown in Figure \ref{fig: LQR}. We represent the LQR state vector as $\mathbf{x} = \left[x, \dot{x}, \theta, \dot{\theta}\right]^T$. $\theta$ represents the tilt angle. With the robot's pose provided, we can calculate the equivalent mass, inertia, and length:
\begin{gather}
m, I, l = m, I, l_{pendulum}(\alpha_l, \alpha_r, \beta_l, \beta_r, \gamma_l, \gamma_r, \phi_l, \phi_r) 
\end{gather}



\subsubsection{Dynamic Equations Derivation}
For the given model, the Lagrangian equations of motion are given by
\begin{gather}
L = T - V \\
\mathrm{\frac{d}{dt}}\left(\frac{\mathrm{\partial} L}{\mathrm{\partial}\dot{q_i}}\right)-\frac{\mathrm{\partial} L}{\mathrm{\partial} q_i} = Q_i
\end{gather}
where $q_i$ represents each generalized coordinate and $Q_i$ stands for generalized force term. According to Kim \cite{kim2011dynamic}, the dynamic equation of WIPS can be driven as below:
\begin{gather}
(m_{w} + m_{p} + \frac{I_{w}}{R^2})\ddot{x} +m_{p} l_{p}\cos{\theta} \ddot{\theta} - M_{p}l_{p}\sin{\theta}\dot{\theta}^2 = \frac{M}{R}\label{Dynamic1}\\
(m_{p} l_{p}^2+I_{p})\ddot{\theta} + m_p\ddot{x}l_p\cos{\theta} - m_{p}gl_{p}\sin{\theta} = -M\label{Dynamic2}
\end{gather}
where the subscripts $p$ and $w$ represent pendulum and wheel respectively.

\subsubsection{Stabilizing Control}
To achieve stable sliding motion, a robust stabilization algorithm is required. The target of the algorithm is to make the robot handle external disturbances to keep its equilibrium, using as little time and energy as possible. However, time saving and energy saving are contradictory in most cases. To solve this problem, LQR, an optimal control method for controlling a linear system with minimal cost, is used in the robot. Li, Yang, and Fan \cite{c22} demonstrated that employing an LQR controller effectively ensures strong reliability and robustness in stabilization scenarios involving a two-wheeled inverted pendulum.
\par

The state equation for the LTI (Linear Time-Invariant) discrete system is as follows:
\begin{gather}
\mathbf{x}\left( {k + 1} \right) = \mathbf{A}\mathbf{x}(k) + \mathbf{B}\mathbf{u}(k).
\end{gather}
In the equation, $\mathbf{x}$ is the state vector, $\mathbf{u}$ is the input vector. The control matrices $\mathbf{A}$ and $\mathbf{B}$ define the discrete-time state-space representation of the system, which is derived from the linearization of the robot model (in formula \ref{Dynamic1} and \ref{Dynamic2}) around the state vector $\mathbf{x} = [0, 0, 0, 0]^T$. The quadratic performance index $J$ is denoted as
\begin{gather}
J = {\sum\limits_{k = 0}^{\infty}{\mathbf{x}^{\mathbf{T}}(k)\mathbf{Q}\mathbf{x}(k) + \mathbf{u}^\mathbf{T}(k)\mathbf{R}\mathbf{u}(k)}}
\end{gather}
In the given equation, $\mathbf{Q}$ and $\mathbf{R}$ represents the weighting matrices for the state covariance. The optimal gain matrix $\mathbf{K}$ can be determined by solving the discrete time algebraic Riccati equation. The robot is dynamically adjusted by the following control law, where $\mathbf{u}$ contains $M$ in Figure \ref{fig: LQR} 
\begin{gather}
\mathbf{u} = -\mathbf{K}\mathbf{x}
\end{gather}
where the state vector $\mathbf{x}$ can be derived from the kinematics.

Besides, for primary pose control of the legs, position control is employed, corresponding to the modification of equivalent mass parameters in the inverted pendulum model.

\subsection{Walking: Reinforcement Learning Control}

For robust walking and jumping control of 10-DOF robot, we conduct the PPO algorithm \cite{PPO} in the simulation. PPO is a reinforcement learning algorithm that applies proximal policy optimization to train policies. It ensures stability by constraining the magnitude of policy updates and uses a clipping mechanism to control the extent of updates.

The target function is given by
\begin{gather}
L^{CLIP}(\theta)=\hat{E}_{t}[min(r_{t}(\theta)\bar{A}_{t},clip(r_{t}(\theta),1-\epsilon,1+\epsilon)\bar{A}_{t})]
\end{gather}
The first term of the minimum function is $L^{CLIP}$. The second term, $clip(r_{t}(\theta), 1-\epsilon,1+\epsilon)\bar{A}_{t}$, is used to restrict the range of the probability ratio $r_{t}(\theta)$ by clipping it between $1-\epsilon$ and $1+\epsilon$ (where $\epsilon = 0.2$), ensuring that it does not deviate too far from 1.

\begin{figure*}[t]
      \centering
      \includegraphics[width=0.95\linewidth]{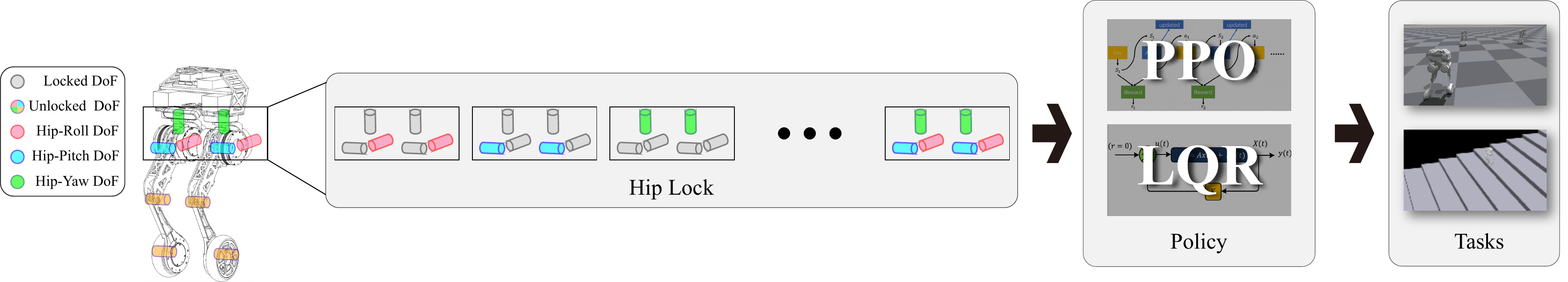}
      \caption{Schematic diagram of the balance experiment.}
      \label{fig: balance experiment schematic}
\end{figure*}

As for the reward, we use a fine-tuned reward function to regulate the robot’s actions, enabling it to walk and jump. We set termination and joint position limits to negative values, with large absolute magnitudes. The purpose is to expedite the robot's convergence to a stable state where it maintains balance and adheres to joint constraints. Taking jumping as an example, we reward the time when its feet are off the ground, the velocity along the positive direction of the Z-axis, and the base height. Meanwhile, rotation and angular velocity are penalized. These lower-value reward functions ensure that the robot's jumping occurs in place.



\section{EXPERIMENTS}\label{section: experiments}

\subsection{Simulation Platform}
We have established a distributed simulation platform comprising three components: 

\begin{enumerate}
    \item Simulator: Simulates environments and robot motion. 
    \item Controller: Executes robot motion control algorithms.
    \item Communicator: Links the controller and simulator via Robot Operating System (ROS). 
\end{enumerate}

This framework decouples simulator, control algorithms, and communication protocols. It facilitates the adaptation of these algorithms for real-world applications.

Drawing from the methodology presented in Legged Gym \cite{leggedgym}, we utilize IsaacGym \cite{isaacgym} for GPU-based parallel training. We use PhysX and apply LQR and PPO algorithms.

\subsection{Experimental Results}
We designed experiments to showcase the advantages of Whleaper in terms of stability (IV-B.1), agility (IV-B.2), and flexibility (IV-B.3).

\subsubsection{Stability}\label{balance}

The experimental design is illustrated in Figure \ref{fig: balance experiment schematic}. Within the simulation, we initially restrict certain DOFs at the robot’s hip to create different configurations of the wheel-legged robot. Subsequently, we attempt to control the robot using either LQR or PPO algorithms and compare their performance in executing specific motion tasks. Among these tasks, testing the walking stability stands out as particularly significant. In this experiment, we employ RL to guide the robot in a forward walking motion while concurrently monitoring the fluctuations in base attitude. 

\begin{figure}[H]
    \centering
     \includegraphics[width=0.95\linewidth]{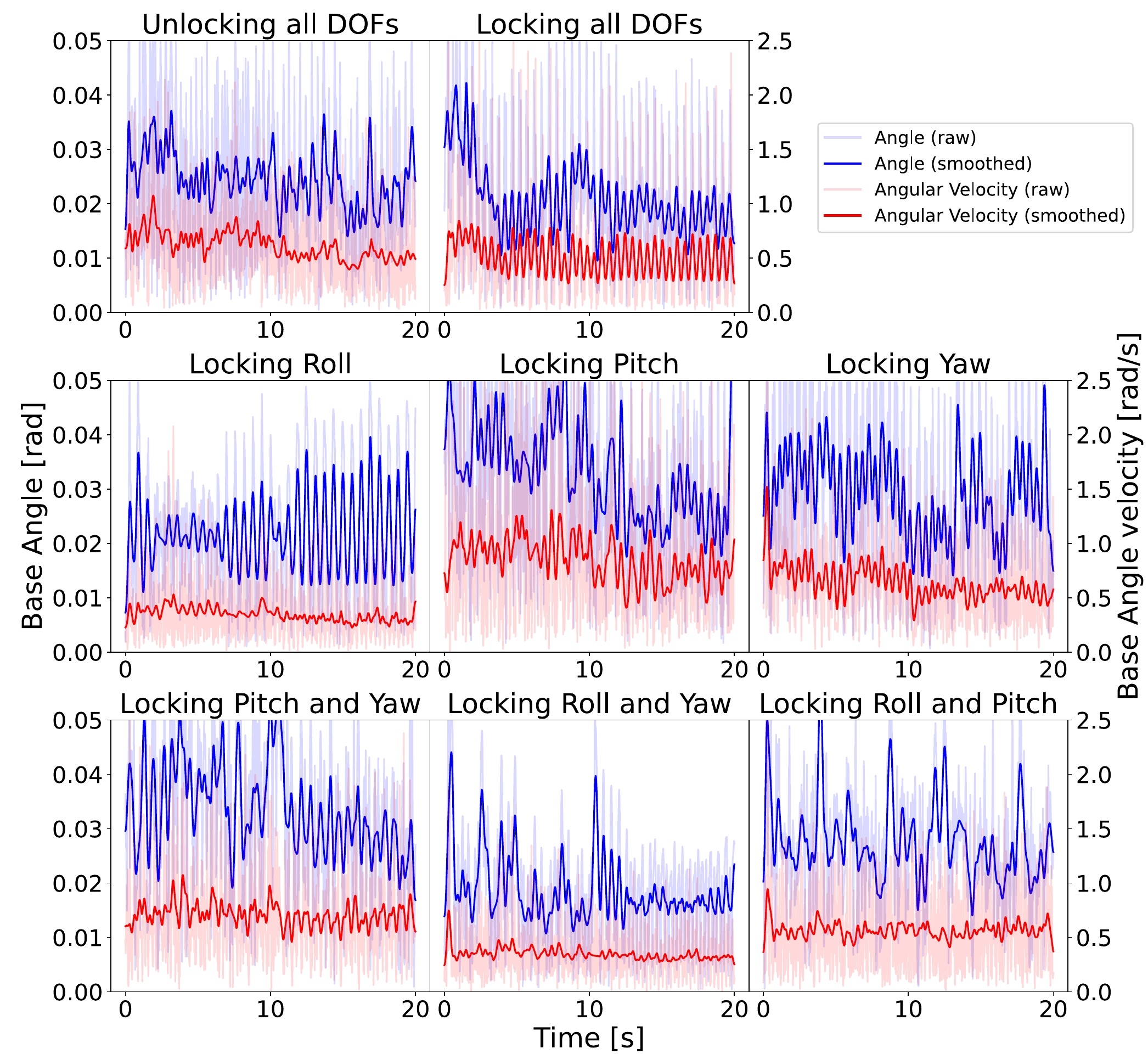}
    \caption{Base variation under different DOFs in the balance experiment. The dark curve represents the actual data, while the light curve indicates the data after Gaussian smoothing. }
    \label{fig: balance experiment data}
\end{figure}

The base deviation depicted in the Figure \ref{fig: balance experiment data} is quantified by the Root Mean Square (RMS) of the Roll and Pitch angles and angular velocities. The result illustrates that as the number of DOFs increases, the variations in base angles and angular velocities tend to decrease. Specifically, the hip-pitch DOF has the most significant impact, with the other two DOFs also exhibiting discernible effects.

With reference to the video, the increase in DOFs enhances the selection of foot placement during walking and expands the robot’s state space, which may improve walking stability. It is noteworthy that while locking the roll and yaw DOFs results in smaller changes in angle and angular velocity, the gait appears slower and less natural according to the video.

\subsubsection{Agility}\label{jumping}
\begin{figure}[!hb]
    \centering
    \subfloat[Twist with 10 DOFs]{\includegraphics[width=0.4\linewidth]{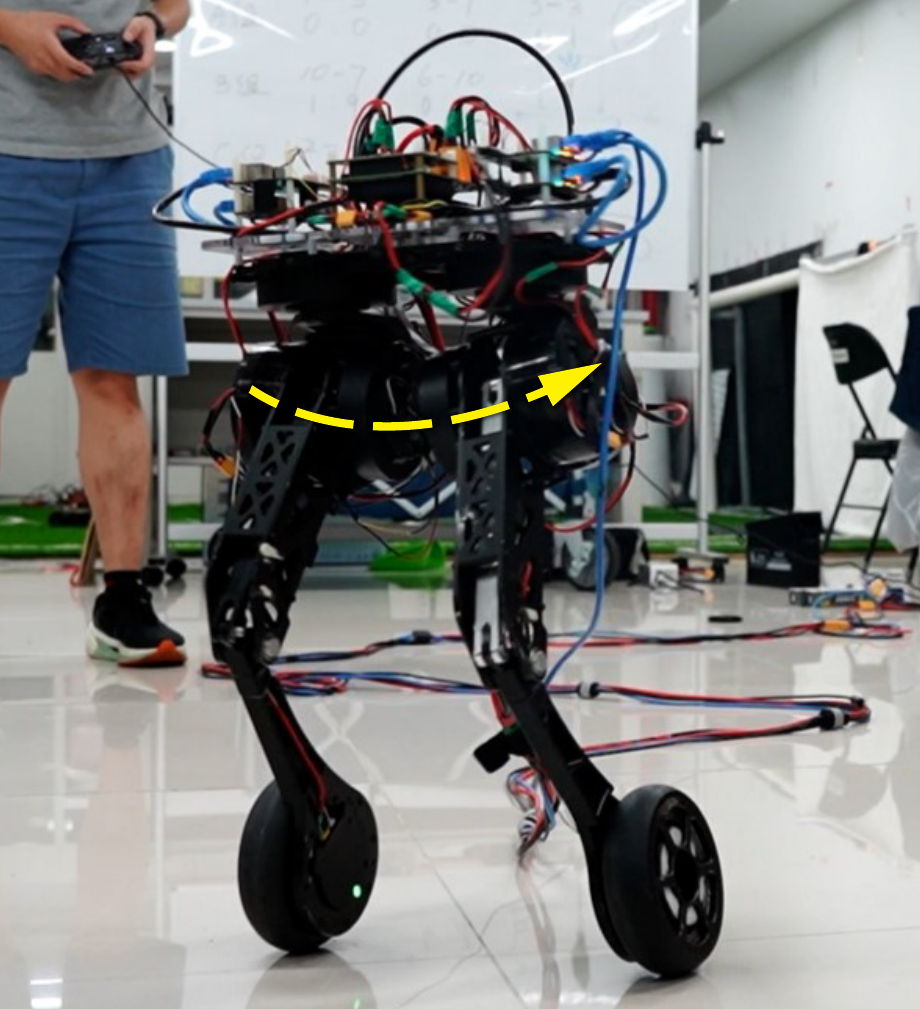}}
     \hspace{0.05\linewidth} 
    \subfloat[Twist with 8 DOFs]{\includegraphics[width=0.4\linewidth]{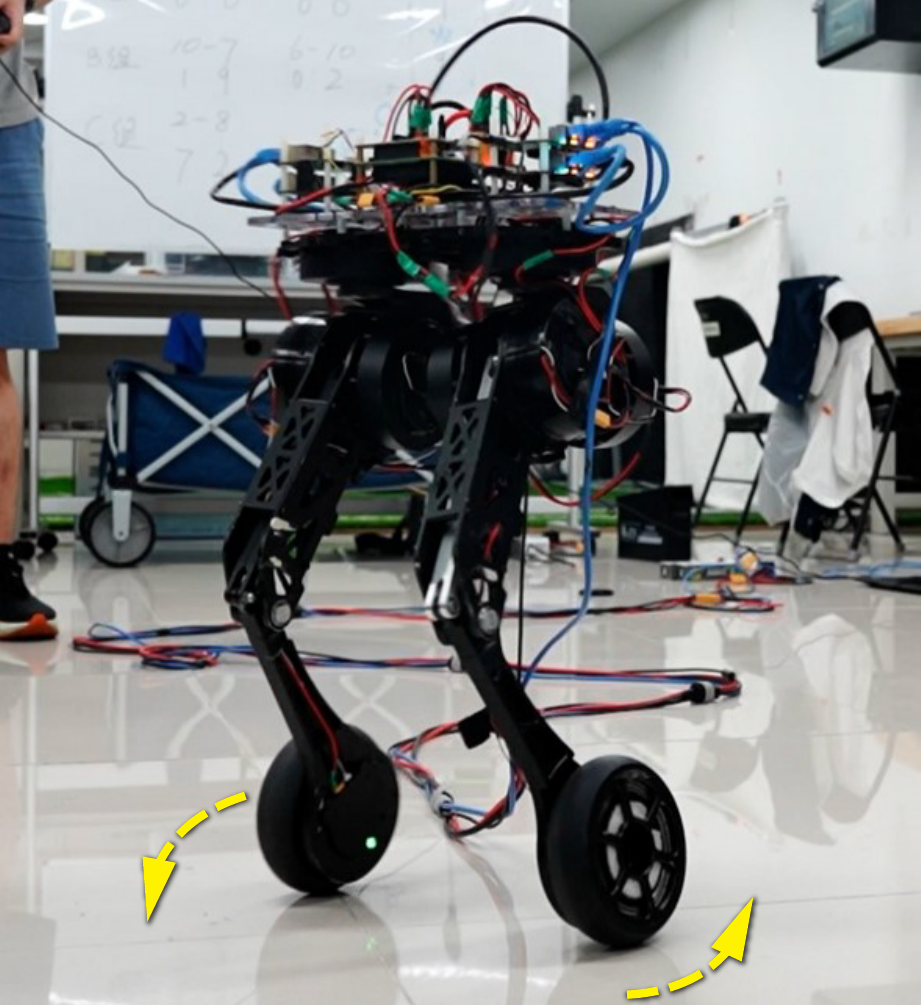}}
    \caption{Photos of twist motion under different DOFs.}
    \label{fig: speed}
\end{figure}

In real-world scenarios, we coordinate multiple joints of Whleaper by LQR control, for achieving yaw angle variation in place, referred to as 'twist'. As shown in Figure \ref{fig: speed} and the accompanying video, Whleaper with 10 DOFs exhibits smoother and faster yaw angle changes compared to configurations with 8 DOFs. This improvement stems from the 10-DOF robot's ability to adjust base rotation primarily through changes in leg posture with minimal wheel movements. In contrast, robots with 6 DOFs (e.g., Ascento) rely solely on differential wheel speeds for rotation. Thus, the addition of hip joints can significantly enhance the execution of specific maneuvers.

\subsubsection{Flexibility}\label{agility}
\begin{figure}[!htbp]
\centering
\subfloat[Forward walking]{\includegraphics[width=0.48\linewidth]{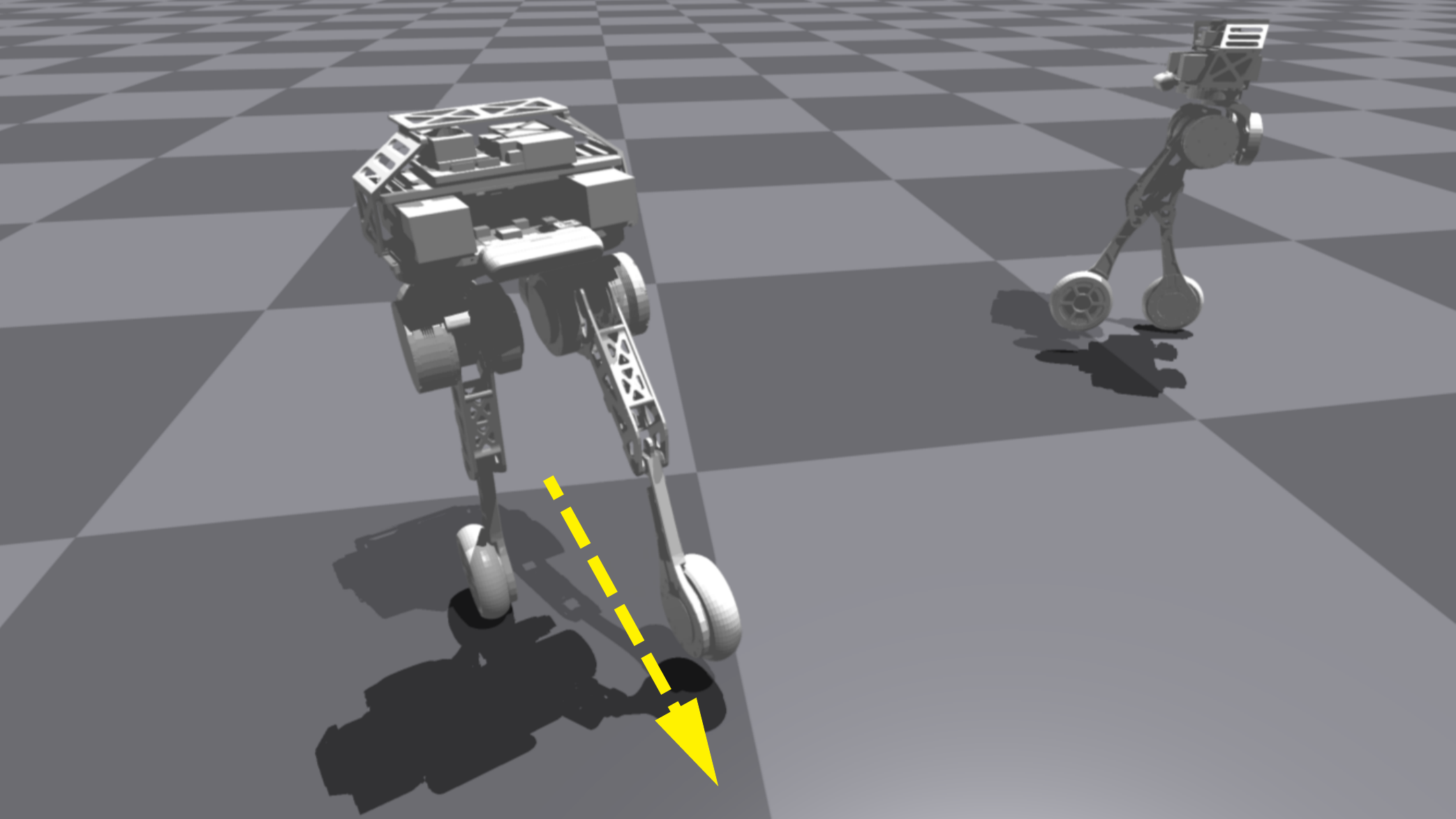}\label{photos-walking}}
\hfill
\subfloat[Jumping]{\includegraphics[width=0.48\linewidth]{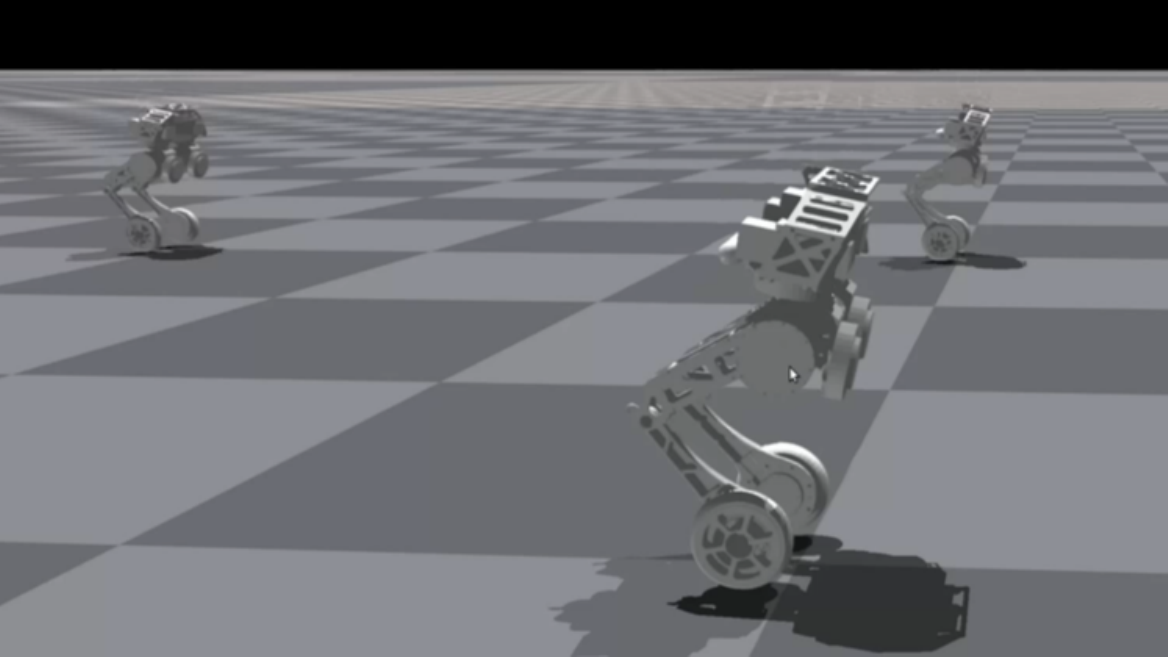}\label{photos-jumping}}
\\[-2ex]
\subfloat[Lateral walking]{\includegraphics[width=0.48\linewidth]{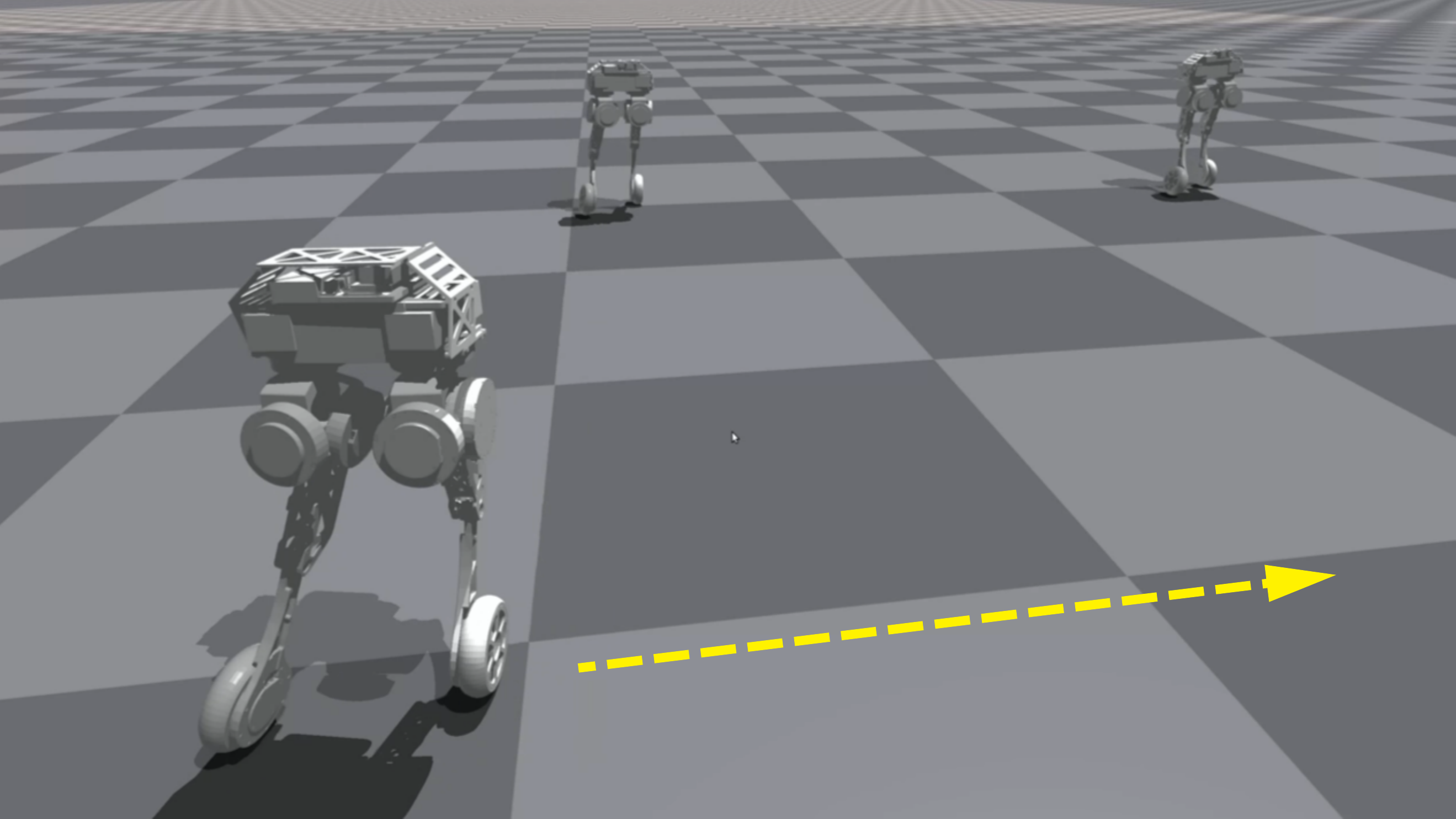}\label{photos-lateral_walking}}
\hfill
\subfloat[Obstacle avoidance sliding]{\includegraphics[width=0.48\linewidth]{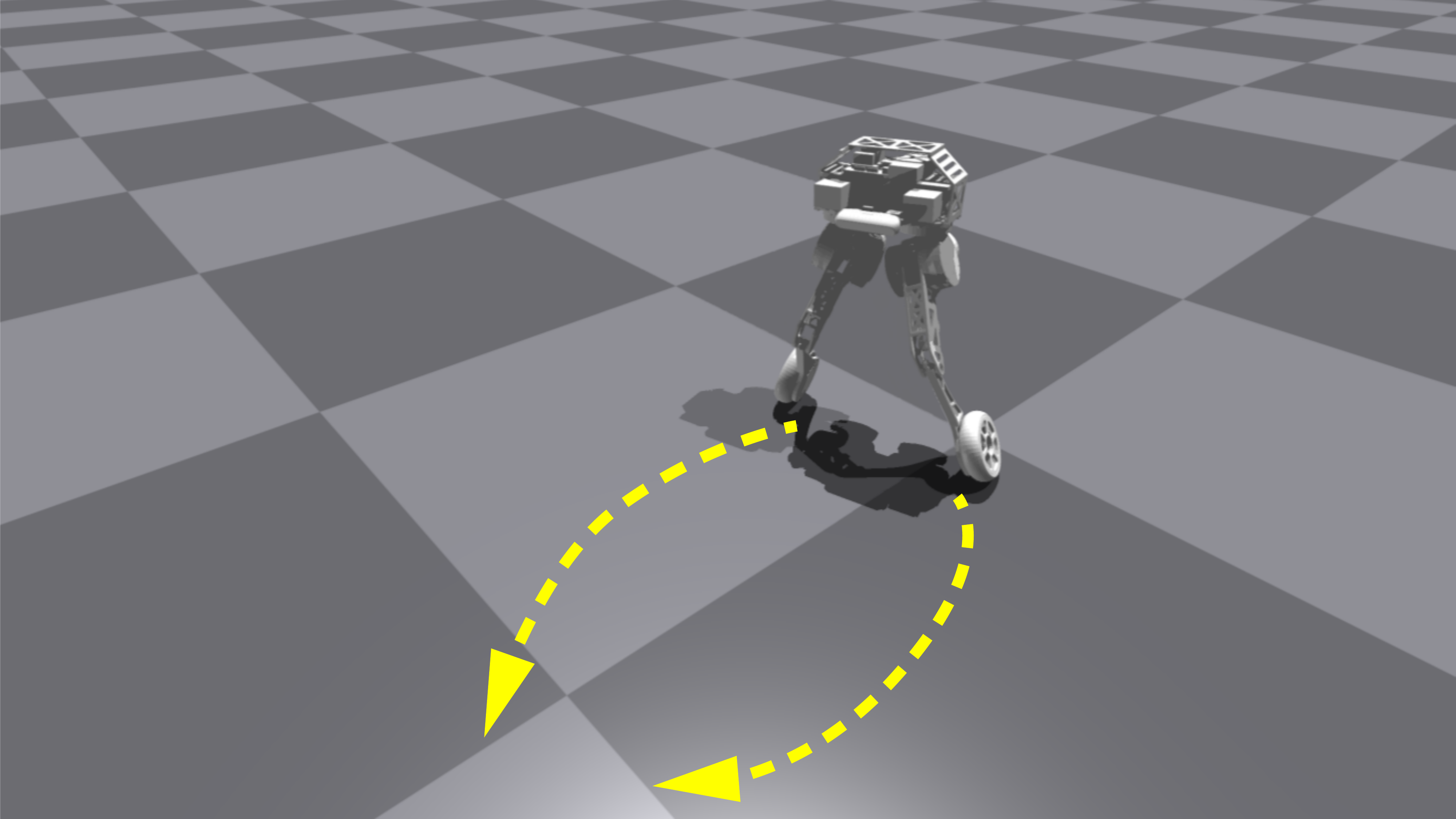}\label{photos-obstacle_avoidance_sliding}}
\\[-2ex]
\subfloat[Skate-like turn]{\includegraphics[width=0.48\linewidth]{fig/Skate-like_Turning.pdf}\label{photos-skate-like_turn}}
\hfill
\subfloat[Knee swing]{\includegraphics[width=0.48\linewidth]{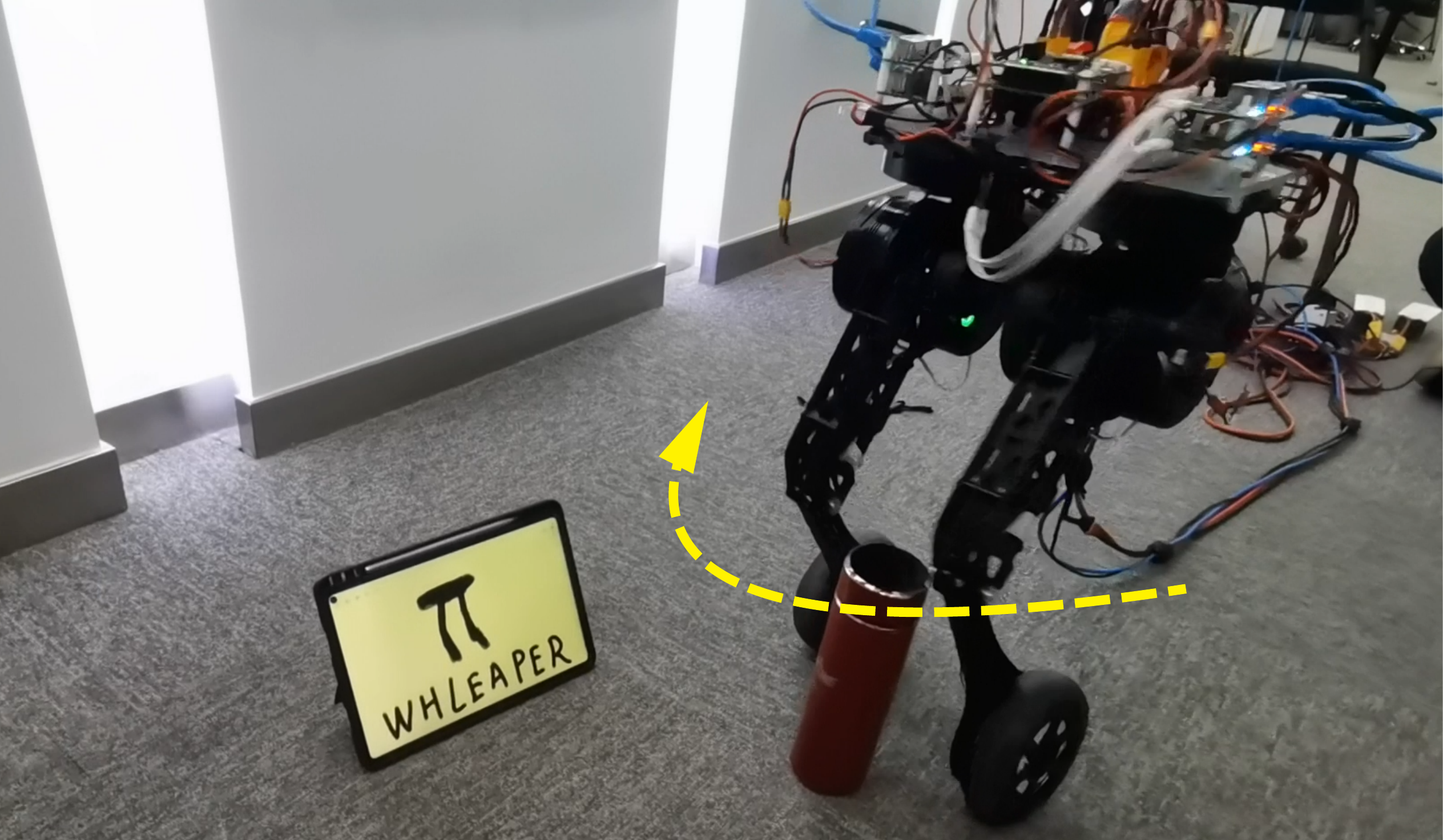}\label{photos-knee_swing}}
\\[-2ex]
\subfloat[Squat]{\includegraphics[width=0.48\linewidth]{fig/MountMoving.pdf}\label{photos-squat}}
\hfill
\subfloat[Crossing barriers]{\includegraphics[width=0.48\linewidth]{fig/Crossing_Barriers.pdf}\label{photos-croosing_barriers}}
\\[-2ex]
\subfloat[Standing on a slope]{\includegraphics[width=0.48\linewidth]{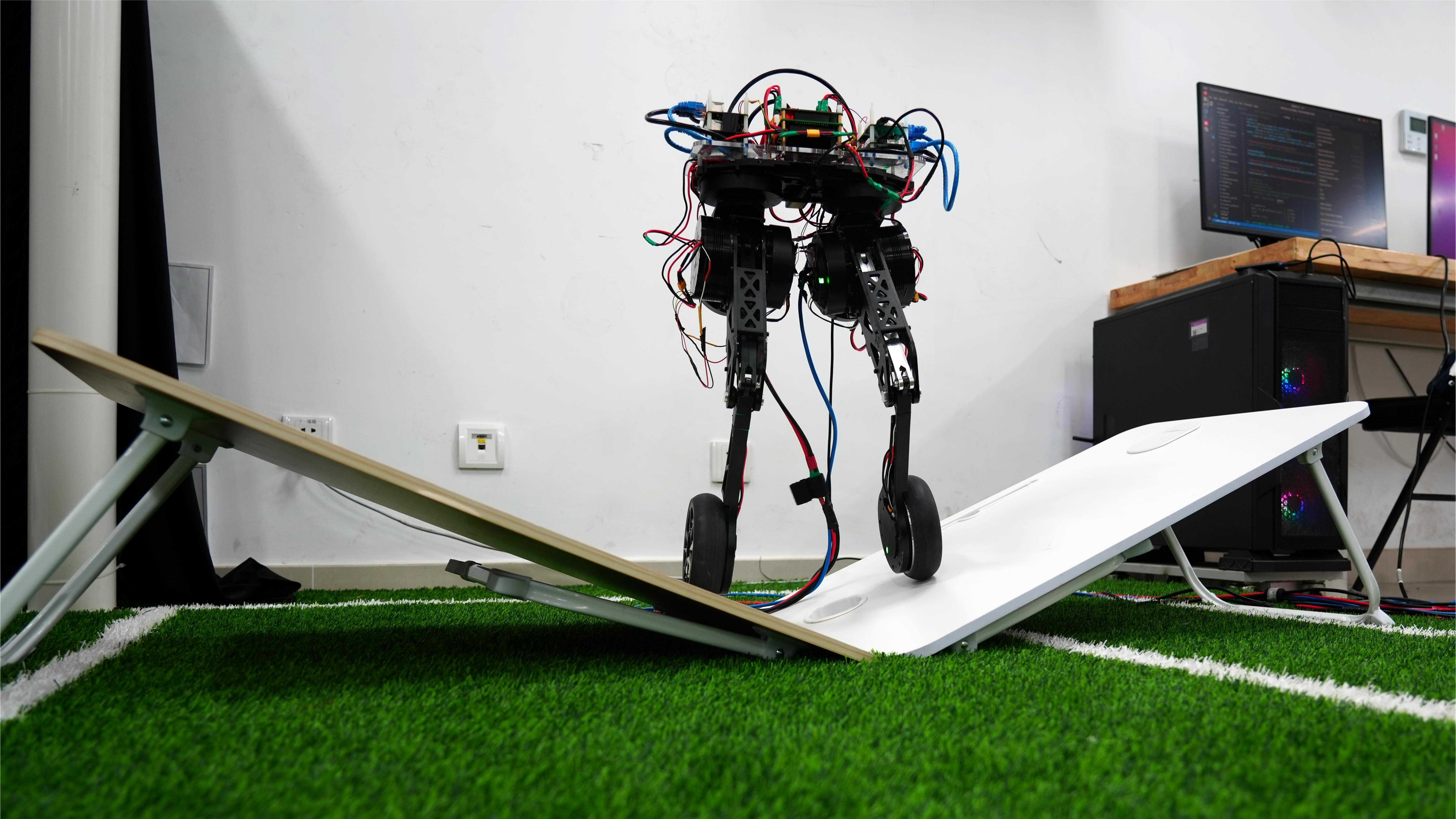}\label{photos-standing_on_slope}}
\hfill
\subfloat[Sliding along a S-path]{\includegraphics[width=0.48\linewidth]{fig/Sliding_along_a_S-path.pdf}\label{photos-sliding_along_a_S-path}}
\\[-1ex]
\caption{Photos of the robot in a series of motions, in simulation and reality.}
\label{fig:photos}
\end{figure}

As depicted in Figures \ref{fig:photos}, Whleaper demonstrates various human-like actions, highlighting its high-DOF advantage.

In narrow indoor environments, in addition to reaching destinations through a combination of straight and turning movements, it may be necessary to control the position and posture of the robot's foothold to avoid obstacles. Therefore, for the two modes of movement, walking and sliding, we conducted tests using unconventional trajectories respectively. Lateral walking, as shown in Figure \ref{fig:photos}\subref{photos-lateral_walking}, requires coordinated motion of the three hip joints, a capability not achievable by existing wheel-legged robots. Obstacle avoidance sliding, as shown in Figure \ref{fig:photos}\subref{photos-obstacle_avoidance_sliding}, utilizes the hip-roll joint to bifurcate the legs while simultaneously using the hip-yaw joint to alter the wheel's heading. This coordinated motion helps to reduce ground resistance during leg bifurcation. Therefore, the high DOFs at the hip of Whleaper is necessary for precisely controlling the wheel trajectory.

Furthermore, movements like skate-like turn (Figure \ref{fig:photos}\subref{photos-skate-like_turn}) and knee swing (Figure \ref{fig:photos}\subref{photos-knee_swing}) also require all of the hip joints, demonstrating Whleaper's advantages over 6/8-DOF robots.

It's worth noting that, besides flexibility in walking and sliding, in a simulated environment, by altering policies, we can freely switch the robot between LQR-controlled sliding and RL-controlled walking. This motion switching enables the robot to adopt different motion modes when dealing with various scenarios, further enhancing the flexibility of wheel-legged robots. Watching the video provides a comprehensive understanding of the robot's capabilities.

\section{CONCLUSIONS}\label{section: conclusions}
This paper introduces Whleaper, a bipedal wheeled robot with 10 DOFs. Whleaper is equipped with a flexible and robust mechanical structure that enables it to perform a range of tasks, such as walking, jumping, and obstacle avoidance sliding. The robot's versatile design allows for seamless transitions between different modes of locomotion, making it suitable for diverse applications. 

We have successfully implemented LQR and reinforcement learning control in the simulation. It showcases the enhanced stability, agility, and flexibility of the wheel-legged robot by increasing hip DOFs. However, we only utilized LQR in limited scope of real-world experiments. Bridging the sim2real gap requires additional effort. In the future, we aim to transfer the reinforcement learning methods into real-world applications, and implement control strategies such as Model Predictive Control (MPC) and Whole-Body Control (WBC) to improve the precision and reliability of wheel-leg cooperative control. 


\section*{ACKNOWLEDGMENT}

This work was supported by the Intelligent Systems and Robotics Lab (ISR) at IIIS, Tsinghua University (THU). We thank Merlin, Xiang Zhu, Zhi Zhang, Qiuhu Liu, and the ISR Lab team for helps in system setup and experiments.

Gratitude goes to Prof. Huichan Zhao, Prof. Chuxiong Hu, Prof. Mingguo Zhao, and colleagues from Weiyang College, Mechanical, Electrical, and Automation departments, along with Yang Cheng and Xuguang Dong, for their support.

Finally, we sincerely thank to the $41^{st}$ THU Challenge Cup Match, for inspiring our dedication to robotics research.


\bibliographystyle{IEEEtran}
\bibliography{reference.bib}
\end{document}